\pdfoutput=1

\documentclass[11pt]{article}

\usepackage[final]{acl}

\usepackage{times}
\usepackage{latexsym}

\usepackage[T1]{fontenc}

\usepackage[utf8]{inputenc}

\usepackage{microtype}

\usepackage{inconsolata}

\usepackage{graphicx}
\usepackage{xspace}
\usepackage{subcaption}
\usepackage{tabularx,colortbl}
\usepackage{booktabs}
\usepackage{multirow}
\usepackage{amssymb}

\title{Investigating Subjective Factors of Argument Strength: Storytelling, Emotions, and Hedging}

\author{Carlotta Quensel \\
  Leibniz University Hannover \\
  \texttt{c.quensel@ai.uni-hannover.de} \\\And
  Neele Falk \\
  University of Stuttgart \\
  \texttt{neele.falk@ims.uni-stuttgart.de} \\\AND
  Gabriella Lapesa \\
  Leibniz Institute for the Social Sciences - GESIS \&  Heinrich Heine University  Düsseldorf \\
  \texttt{gabriella.lapesa@gesis.de} \\}

\newcommand{\corpusname}[1]{{\small \textsc{#1}}}
\newcommand{\ibmargq}{\corpusname{IBM ArgQ}\xspace}
\newcommand{\cmv}{\corpusname{Cornell CMV}\xspace}
\newcommand{\crowd}{\corpusname{Crowd-enVENT}\xspace}
\newcommand{\F}{F$_1$\xspace}
\newcommand{\bspubtag}{
    \vspace*{-19cm}\hspace*{-0.5cm}
    {\fontsize{6}{8}\selectfont%
        \renewcommand{\arraystretch}{0.9}
        \begin{tabular}{l}
            Accepted to \\
            12th Workshop on Argument Mining (ArgMining) 2025
        \end{tabular}
}}

\begin{document}
\maketitle
\begin{abstract}
In assessing argument strength, the notions of what makes a good argument are manifold. With the broader trend towards treating subjectivity as an asset and not a problem in NLP, new dimensions of argument quality are studied. Although studies on individual subjective features like personal stories exist, there is a lack of large-scale analyses of the relation between these features and argument strength. 
To address this gap, we conduct regression analysis to quantify the impact of subjective factors -- emotions, storytelling, and hedging -- on two standard datasets annotated for objective argument quality and subjective persuasion. As such, our contribution is twofold: at the level of contributed resources, as there are no datasets annotated with all studied dimensions, this work compares and evaluates automated annotation methods for each subjective feature. At the level of novel insights, our regression analysis uncovers different patterns of impact of subjective features on the two facets of argument strength encoded in the datasets. Our results show that storytelling and hedging have contrasting effects on objective and subjective argument quality, while the influence of emotions depends on their rhetoric utilization rather than the domain.
\end{abstract}

\bspubtag
\vspace*{18cm}\hspace*{0.5cm}

\section{Introduction}
Argument Mining describes the field of detecting arguments and their components, i.e., claims and their premises, and analyzing relationships like support and attack between those \citep{lawrence-reed-2019-argument}. This notion of argumentation as primarily reason-giving, paired with the prominent domains of academic writing, student essays, or professional debate, necessitating objectivity for judging and automatic essay scoring, led to a narrow conceptualization of argument quality. Quality assessment, as emerged from argument mining, observes \textbf{objective aspects} such as clarity and argument organization \citep{persing-etal-2010-modeling}, use of evidence \cite{rahimi-2014-evidence}, or a combination of those \citep{ong-etal-2014-ontology}.
In the past years, however, a clear need for a shift towards a more subjective notion of argument quality has emerged, driven by the entry of laypeople into the debate space through online forums and citizen participation programs, as well as insights contending the link between objective quality and persuasive strength \cite{benlamine2017persuasive-emotions}. This paper contributes to a better empirical understanding of the impact of subjectivity on argument quality. 

\begin{table}
    \centering\small
    \setlength\tabcolsep{2pt}
    \begin{tabularx}{\linewidth}{|X|}
    \hline
Sports offer a lot more than you'd think\dots\\
1) It gives children a sense of being a part of something (crucial for kids without stable families)\\
2) Sports are a GREAT source of exercise\\
{[}\dots{]} There's \textbf{many more} reasons but this is all I \textbf{can think} of for now. As for my own experiences, baseball and football has helped me come out of my shell and meet some of the best people I've ever met in my life. I \textbf{don't know} where I'd be without these sports.\hspace*{\fill}($\Delta1$, \textit{joy, story}, $\varnothing$\textit{hedges}=0.051)\\\hline
\end{tabularx}
\caption{Annotated \cmv instance with positive labels listed at the end and boldened hedge terms.}\label{tab:intro-ex}
\end{table}
More specifically, we focus on three subjective features, namely emotions, storytelling (personal and/or anecdotal narratives), and hedging (terms marking uncertainty, e.g., \textit{probably, I think, likely}). While these aspects have already been investigated individually, i.e., in works investigating the use of personal narratives in argumentation \citep{falk-lapesa-storytelling}, emotional progression \citep{benlamine2017persuasive-emotions}, or human values \citep{kiesel-etal-2022-identifying}, the crucial element of novelty of this work is the fact that we consider the (joint) impact of such subjective features on argument strength as opposed to previous work that considers them in isolation. Table \ref{tab:intro-ex} shows an argument appealing to \textit{joyful} emotions and personal experiences, while recognizing knowledge gaps. The argument originates from the online forum \textit{r/ChangeMyView}, where the user was successful ($\Delta1$) in the forum's goal of persuading the discussion's initiator, showing the importance of investigating these features and their impact on argument quality more rigorously.

Toward this end, we carry out a parallel analysis on two datasets containing argument quality annotations which approximate the diverging conceptualizations of argument strength related to the function of argumentations: for the reason-giving function we selected \ibmargq \citep{toledo-etal-argument-quality}, whose annotations encode \textbf{objective argument quality}; for the persuasion function, we selected the \cmv dataset \citep{tan-changemyview} aggregated from the previously mentioned \textit{r/ChangeMyView} forum, whose metadata (i.e., the presence of a delta indicating that the originator of a discussion changed their opinion following a specific answer) encode \textbf{individualized persuasion}. Differing not only in collection method, domain, argument length, and annotation procedure, these two datasets also lend themselves as the perfect pairing for a contrastive analysis of the impact of subjective features. 

Our work proceeds in two steps. As a first step, we automatically enrich the two datasets with one annotation layer per subjective feature. To this end, we compare and evaluate alternative annotation methods (cf. Sec.~\ref{sec:anno}) and reflect on their properties and suitability for our domains of interest. In our second step (Sec.~\ref{sec:analysis}), we address the main research goal of the paper: the impact of subjective features on argument strength. We employ regression analysis and address two research questions: \textbf{RQ1:} Do subjective features impact argument strength? \textbf{RQ2:} Do the patterns of their impact differ in the comparison between objective argument quality and individualized persuasion? 

The contributions of our work are accordingly twofold. At the level of novel insights on the phenomenon of argument quality, our work is the first one that targets the \textit{joint} impact of storytelling, emotion, and hedging on argument quality. %
At the level of contributed resources, we release and share with the community the datasets with the new annotation: this will enable further research on the interplay of these phenomena.\footnote{Data and code are available at: \url{https://github.com/CarlottaQuensel/subjective-argument-strength}}

\section{Related Works}\label{sec:related}
\subsection{Argument Strength}
The question of what makes a good argument has been studied since Aristotle (\citeyear{aristotle2007rhetoric}), who devised three main strategies of \textit{ethos} or appeal to authority (of experience or persona), \textit{pathos} or appeal to emotions, and \textit{logos} or appeal to logic. The latter strategy maps onto the notion of argumentation as \textit{reason-giving}, which has historically been favored in research.
In both computational argumentation and the social sciences, a primary view of argumentation as a rational, somewhat mechanistic process of finding the objectively best claim through a combination of premises and evidence narrowed the notion of argument quality into one of successful \textit{logos} rhetoric. In the predominant domains of student essays and professional debate, this is necessary, but limits the features and dimensions investigated in relation to argument quality to the objective and logical. As such, there are several investigations into clarity, use of evidence, or organization \citep{persing-etal-2010-modeling, persing-ng-2013-modeling, rahimi-2014-evidence}, with multiple argument quality corpora using corresponding definitions: ease of understanding \citep{swanson-etal-2015-argument} or the general suitability as part of a larger thesis \citep{toledo-etal-argument-quality,gretzFCTLAS20}. These datasets are usually annotated by merging crowdsourcing labels, which further affirms the notion of argument quality as an, if not explicitly objective, then explicitly universal measure.  

The inclusion of user-generated arguments in informal online settings shifted this focus at a similar time as the \textit{affective turn} in the political sciences \citep{hoggett2012}, reorienting argument strength notions toward the persuasion function of argumentation as well as \textit{ethos} and \textit{pathos} strategies. This shift produced multiple studies of features related to \textit{ethos}, mainly codifying meta-information such as %
prior beliefs, personal characteristics, and human values \citep{lukin-etal-2017-argument,al-khatib-etal-2017-patterns,kiesel-etal-2022-identifying}, or, only recently, personal narratives as a form of non-traditional expertise \citep{falk-lapesa-storytelling,falk-lapesa-2023-storyarg}. \textit{Ethos}-related works mainly looked at emotional appeal \citep{benlamine2017persuasive-emotions} or fallacious emotions \citep{ziegenbein-etal-2023-modeling}. While multiply new datasets were published in parallel to these studies, targeting \textit{convincingness} and \textit{persuasion} \citep{habernal-gurevych-2016-argument,simpson-gurevych-2018-finding,gleize-etal-2019-convinced}, or aiming to codify all existing dimensions of argument quality into a cohesive taxonomy and annotation hierarchy \citep{wachsmuth-etal-2017-computational,ng-etal-2020-creating}, many of these datasets similarly encode argument quality as a universal average of multiple crowdworkers, thus blurring the distinction between objective and subjective dimensions.

Thus, a gap becomes apparent in the understanding of features relating to \textit{ethos} and \textit{pathos}, such as the establishment of personal authority through \textit{storytelling} or \textit{hedging} and the direct investigation of individual \textit{emotions}. Though these three features hold promise for argument assessment, they are largely understudied in Computational Argumentation.
\subsection{Subjective Argument Features}
    
\paragraph{Storytelling} 
Research on personal testimonies or \textit{storytelling} originates from the field of deliberative research, where it has long been recognized as a tool to convey empathy and lived experience \citep{black2008-storytelling,black2013-storytelling,Esau-2018}. By establishing personal expertise, personal narratives aid in the construction of \textit{ethos}, though \citet{maia-cal-bargas-crepalde-2020} show how narratives enrich debates in public hearings, incorporating \textit{logos} and \textit{pathos} in complex ways. Thus, storytelling serves as an alternative evidency type for non-experts and allows for disagreements without direct conflicts of facts. These observations, however, stem from small case studies and in Computational Argumentation, storytelling only recently gained attention. \citet{el-baff-etal-2020-analyzing} included the number of anecdote sentences in news editorials, but do not address the feature separately.
\citet{falk-lapesa-storytelling,falk-lapesa-2023-storyarg} consolidate multiple small social science datasets to allow for computational investigations of the phenomenon and argue that integrating personal narratives into argument mining helps include voices often excluded by logos-centric models. Their exploratory findings suggest that storytelling may positively correlate with several quality dimensions in an annotated corpus, but the effects on overall argument quality remain underexplored in a large scale or systematic analysis.

\paragraph{Emotion} There are multiple investigations into the impact of emotions on arguments, though investigations of multiple discrete emotions are scant, small, and very recent. Most Computational Argumentation approaches collapse \textit{emotion} and \textit{emotional appeal} into one feature modeled as stance, polarity (e.g., \citealp{grosse2015integrating,stede2020automatic,el-baff-etal-2020-analyzing}), intensity, or the general presence of any emotion \citep{fromm-etal-argument-emotions}. Further, \textit{emotional appeal} is historically seen as a fallacy in rational discourse, leading to multiple works investigating emotions as a negative feature (e.g., toxic emotions, \citealp{ziegenbein-etal-2023-modeling}). The argument quality taxonomy and dataset by \citet{wachsmuth-etal-2017-computational} also includes emotional appeal in its 15 labels. In the deliberative field, \citet{maia2020emotional} observe \textit{anger}, \textit{fear}, \textit{indignation} and \textit{compassion} in political discussions, showing how these emotions are distributed unevenly between different argument directions. \citet{benlamine2015emotions,benlamine2017persuasive-emotions} showed the link between emotions and argumentation behavior and found that, from Aristotle's rhetoric strategies, emotional appeal (\textit{pathos}) is most persuasive. Only recently, the first (to our knowledge) small dataset of 1031 German arguments annotated for convincingness and 10 discrete emotions was released by \citet{greschner2024fearfulfalconsangryllamas}. Despite this encouraging first step, there are, however, neither other (English) datasets nor large-scale analysis of emotions and argument strength available as of yet.

\paragraph{Hedging} is one of multiple strategies to verbalize the epistemic modality of a proposition \citep{lyons1977}, i.e., convey its degree of certainty (\textit{likely}) or speaker-commitment (\textit{according to \dots}). 
In academic writing, it reflects the precision and caution of the scientific inquiry process, anticipating objections and gaining community acceptance \citep{Hyland1998HedgingIS,Martin2022-dg}. In the fields of medicine and law, hedging serves as a professional face-saver, to build rapport with colleagues, patients, or a jury, and to avoid misinterpretation, thus enhancing speaker credibility \citep{bryant-norman-1979-uncertainty,prince1982physicians,Zaitseva_2023}.
Informally, hedging is investigated as a strategy of politeness and positive self-image \citep{ardissono1999politeness}, %
and as a cooperative strategy to indicate openness to corrections and change \citep{vasilieva2004gender,jordan-2012-uncertainty}.

Thus, with the rhetoric strategy of \textit{ethos} encompassing recognized expertise, hedging is directly tied to this strategy. Wielded purposely, it appeals to the honest conduct and credibility of a speaker, similar to storytelling, although apparent uncertainty may just as well hamper recognized expertise. Despite this relevance, hedging is rarely studied in Computational Argumentation:
Existing works link hedging to debaters' improvement \citep{luu-etal-2019-measuring}, predict persuasiveness with paraverbal hesitation cues \citep{chatterjee-etal-2014-verbal} or modal verbs \citep{wei-etal-2016-post}, but few address the size and direction of any observed effects. \citet{habernal-gurevych-2017-argumentation} show an uneven distribution of hedges skewed toward constructive, nonpolarized discussions. Only \citet{tan-changemyview} directly observe a positive effect on persuasiveness. The mixed findings highlight a gap: Given its surface-level detectability and interpretive flexibility, hedging is a promising but overlooked feature for capturing subjective argument quality. Hedging might enhance argument strength by boosting credibility, or weaken it by implying doubt -- yet no systematic study explores this trade-off.

\section{Data}\label{sec:data}

Investigating the link between argument strength and the subjective features of storytelling, emotions, and hedging requires argument data that is annotated not only for argument strength but also for each of these features. As there is currently no such dataset available, a suitable corpus must be aggregated automatically. Multiple corpora are suitable as a base dataset that includes a gold annotation for the target variable (DV) of argument strength. To approximate the diverging conceptualizations of argument strength explicated above, we chose two datasets that differ in collection method, domain, argument length, and annotation procedure, categorized below as objective argument quality and individualized persuasion.

\paragraph{Objective argument quality} \textsc{ibm-ArgQ} 5.3k \cite{toledo-etal-argument-quality} consists of 5.3k short, stand-alone arguments generated at formal debate events by debate club members of varying skill levels and the general audience. Participants were asked to produce short arguments (max. 36 words) after seeing a professional example argument and choosing one of 11 controversial topics, such as privacy laws, gambling, or vegetarianism with two opposing stances, e.g., \textit{We should adopt vegetarianism} and \textit{We should abandon vegetarianism}. Participants were advised to keep arguments impersonal to avoid privacy concerns in the final dataset.\\
The argument strength annotation is an average of binary crowd judgments: for each argument, 15-17 annotators judged its adequacy as part of a debate speech,\footnote{`\textit{Disregarding your own opinion on the topic, would you recommend a friend preparing a speech supporting/contesting the topic to use this argument as is in the speech?'}} which was averaged for the final score to model the ratio of positive judgments. This procedure attests to a rather unspecific conceptualization of generalized `overall' argument strength, as the annotators must employ their own concept and hierarchy of relevant features, e.g., topic relevance, linguistic clarity, or sound rhetoric, and the single binary judgment paired with the averaging makes reconstruction of these features impossible. %
As such, while the utilized notion of argument strength is not explicitly stated `objective', the domain, style, and annotation process of \textsc{ibm-ArgQ} 5.3k invoke an argument strength conceptualization in line with the traditional \textit{logos} focus of the argument mining field, by removing subjective context and aggregating judgements to approximate a generalized, universal, and thus more objective, argument quality score.
Thus, in the following analysis, this dataset is referred to as \ibmargq and represents argument strength as conceptualized by the traditional argument mining field.

\paragraph{Individualized persuasion} \cmv was aggregated by \citet{tan-changemyview} from 11567 comments posted to the Reddit forum \textit{ChangeMyView}\footnote{\url{https://www.reddit.com/r/changemyview/}} between January 2013 and August 2015, where users state their viewpoint with detailed background on their thought process to engage in constructive discussion that aims at changing their view. Thus, in one comment thread, multiple users argue against the same position until the original poster (OP) awards a \textit{delta point} ($\Delta$) to one or more answers that persuade them. The unique setup of the forum provides an inherent annotation and ensures data quality, with the delta point system that denotes the OP's persuasion and posting guidelines that are actively moderated by volunteers both for civility and for maintaining a constructive discussion in which comments must advance the conversation and decisions for delta points must be explained. The resulting label stands in contrast to the score of \ibmargq, as it encodes the subjective change in opinion of one person from a specific argument, in the context of a mutual discussion and multiple alternative arguments. %
The domain properties further make for much longer texts, sometimes containing multiple premises and stances forming a rhetoric argumentative sequence or direct quotes from the OP, which are addressed point by point. %
In the dataset used here (henceforth \cmv), the posts are structured as contrasting pairs of comments addressing the same OP, one with and one without a delta point, making for a balanced distribution of the binary persuasiveness label.

Given all the above differences between \ibmargq and \cmv, it is apparent that the two datasets conceptualize arguments as well as argument strength in very different ways. Although the number of differences disallows a comparison of pure argument strength conceptualization without any confounding factors, the inclusion of both corpora in the investigation covers idiosyncrasies across the spectrum of the argument mining field on what argument strength means. Tab.~\ref{tab:argument-example} shows examples from both datasets%
. To illustrate the diverging concepts, in the following analysis, argument strength is called \textit{quality} when investigating \ibmargq and \textit{persuasiveness} for \cmv. 

\section{Automatic Annotation of Subjective Features}\label{sec:anno}

As the two datasets do not have annotations for the investigated features, it is necessary to enrich the datasets with the corresponding annotation layers as a first step. %
Thus, an automated annotation model is devised for each of the three features. In what follows, we describe the computational methods we used to achieve this goal separately for each feature. For storytelling and emotions, an ensemble consisting of ten transformer-based classifiers is trained on annotated data. %
As hedging is a surface feature dependent on individual terms, it is annotated using a simple rule-based algorithm.
The following sections \ref{sec:data-story}, \ref{sec:data-emotion}, and \ref{sec:data-hedging} elaborate on the annotation process of each feature and the resulting statistics on the two argument datasets.
\subsection{Storytelling}\label{sec:data-story}
\paragraph{Training Data} As most storytelling research is comprised of small case studies from the political sciences, we combine multiple datasets from different sources following the approach of \citet{falk-lapesa-storytelling}.
\citet{falk-lapesa-storytelling} use a collection of different datasets and domains covering diverse topics, such as expert-moderated discussions on immigration \citep{gerber-2018-europolis} and consumer debt collection \citep{park-cardie-2018-corpus} and a subset of the online debate forum \textit{r/ChangeMyView}. They consolidate different original annotations indicating wether an argument contains a personal experience or story (1) or not. 
\paragraph{Training Setup} 
We fine-tune RoBERTa transformers \citep{RoBERTa} using a 10-fold cross-validation ensemble, where the full dataset is split into ten parts and ten separate models are trained, each on a different combination of training and validation folds. This ensemble approach is used to produce more robust and stable predictions, as it mitigates variance due to random initialization and training data fluctuations \citep[cf. e.g.,][]{Lakshminarayanan17,MOHAMMED2023757}. For annotation, we apply the majority vote across the ten ensemble models to assign labels to our two target datasets. This setup follows \citet{falk-lapesa-storytelling}, both to replicate the results of the original paper and to harness the identification of mixed-domain training as the most robust configuration for cross-domain generalization, making it most suitable for our \ibmargq data. As their reported same-domain performance for the \textit{ChangeMyView} subset is on par with the mixed-domain classifier, we additionally train a classifier on only this subset to potentially harness this effect for \cmv.
\paragraph{Results} As apparent from the test performance on a heldout dataset (cf. Appendix Tab.~\ref{tab:test-eval})%
, the mixed-domain ensemble prevails over the same-domain classifier, both in terms of performance (\F=.82 vs. \F=.78) and lower variance, which is in line with findings by \citet{falk-lapesa-storytelling}. Otherwise, the performance is on par with the results of the best-performing models of the original experiments \citep{falk-lapesa-storytelling} (\F between .76 and .92), allowing us to continue with the analysis using the \textit{mixed-domain} annotations.
The resulting predictions are, however, very sparse for both corpora (cf. Tab.~\ref{tab:anno-dist}), especially so \ibmargq (0.8\% positive), which can be attributed in part to the unbalanced distribution in the training data (storytelling is the minority class), but more importantly to the brevity and impersonality of \ibmargq instances. To mitigate the sparseness, we follow \citet{Lakshminarayanan17} and interpret the average classification probability as a certainty measure of the binary annotation, thus introducing a richer source of information in the next step.

\begin{table} %
	\centering\small
    \begin{tabular}{lrrrrrr}
        \toprule
         & \multicolumn{3}{c}{\textbf{\ibmargq}} & \multicolumn{3}{c}{\textbf{\cmv}} \\
        \cmidrule(l{.9em}r{.9em}){2-4}
        \cmidrule(l{.9em}r{.9em}){5-7}
         \textbf{Feature} & \textbf{\#} & \textbf{\%} & $\mathbf{\varnothing}$\textbf{P}
         & \textbf{\#} & \textbf{\%} & $\mathbf{\varnothing}$\textbf{P}\\
         \cmidrule{1-1}
        \cmidrule(l{.9em}r{.9em}){2-4}
        \cmidrule(l{.9em}r{.9em}){5-7}
        \textit{anger} & 1,814 & 34.2 & .39 
        & 6,467 & 55.9 & .43 \\
        \textit{boredom} & 116 & 2.2 & .06 
        & 538 & 4.7 & .07 \\
        \textit{disgust} & 2,920 & 55.1 & .54 
        & 5,111 & 44.2 & .37 \\
        \textit{fear} & 347 & 6.6 & .14 
        & 822 & 7.1 & .11 \\
        \textit{guilt/shame} & 107 & 2.0 & .12  
        & 631 & 5.5 & .14 \\
        \textit{joy} & 47 & 0.9 & .07 
        & 208 & 1.8 & .05 \\
        \textit{pride} & 80 & 1.5 & .10 
        & 615 & 5.3 & .12 \\
        \textit{relief} & 64 & 1.2 & .06 
        & 256 & 2.2 & .06 \\
        \textit{sadness} & 175 & 3.3 & .14 
        & 429 & 3.7 & .12 \\
        \textit{surprise} & 0 & 0.0 & .03 
        & 53 & 0.5 & .04 \\
        \textit{trust} & 112 & 2.1 & .07
        & 159 & 1.4 & .04 \\
        \hline
        \textit{storytelling}
        & 45 & 0.8 & .02 
        & 2288 & 19.8 & .22 \\\bottomrule
    \end{tabular}
    \caption{Feature distribution according to the best ensembles for emotion (\textit{masked/aggregated}) and storytelling (\textit{mixed}) on \ibmargq and \cmv, including the number (\#) and ratio (\%) of positive instances, and the corpus-wide average classification probability ($\varnothing$P).}\label{tab:anno-dist}
\end{table}
\subsection{Emotion}\label{sec:data-emotion}
\paragraph{Training Data} As expanded in section \ref{sec:related}, while there are multiple works on \textit{emotionality} (intensity, polarity, etc.) in arguments, there are no works and related datasets modeling discrete emotions in English arguments. As such, our approach has to bridge a gap from the emotion domain to the argument domain. Though recent works showed the capabilities of LLMs in emotion classification \citep[cf. last year's WASSA shared task;][]{maladry-etal-2024-findings}, the zero-shot approach necessitated by our lack of in-domain examples is still outperformed by traditional fine-tuning, given a sufficient amount of high-quality training data \citep{kazakov-etal-2024-petkaz}.
With no emotion-annotated datasets in the argument domain, we selected our training data to best match the register and style of our target data. This precludes both very informal and formal datasets aggregated from Twitter or from novels and news headlines, as well as data collected through emotion-specific emojis, words, hashtags, or forums to avoid surface-level emotion representations with low cross-domain adaptability. Thus, we chose \crowd \citep{troiano-etal-2019-crowdsourcing} as our training data, a crowdsourced dataset of event descriptions for eleven different emotions,\footnote{Generated as, e.g., \textit{I felt fear when: \dots} and analogously.} which allows for an implicit emotion representation. 
\begin{table*}[t]
    \centering
    \begin{tabular}{lcp{14mm}rrp{14mm}r}
        \toprule
         & \multicolumn{3}{c}{\textbf{\ibmargq}} & \multicolumn{3}{c}{\textbf{\cmv}}\\
         \cmidrule(r){2-4}
         \cmidrule(l){5-7} 
         \textbf{IV} & \textit{r}\textsuperscript{2} & \textit{p} & Coef & pseudo-\textit{r}\textsuperscript{2} & \textit{p} & Odds \\
         \hline
        storytelling
        & 0.0047 & 0.0\hspace{\fill}*** & $-$0.182 %
        & 0.0004 & 0.015\hspace{\fill}* & 1.148\\
        \hline
        anger 
        & 0.0011 & 0.009 \hspace{\fill} **& $-$0.026 %
        & 0.0000 & 0.377 & 0.928 \\ %
        boredom 
        & 0.0006 & 0.042 \hfill * & $-$0.050 %
        & 0.0000 & 0.487 & 0.897 \\ %
        disgust 
        & 0.0022 & 0.0 \hspace*{\fill} ***& $-$0.031 %
        & 0.0010 & 0.0 \hspace*{\fill} ***& 0.751 \\ %
        fear 
        & 0.0026 & 0.0 \hspace*{\fill} ***& 0.056 %
        & 0.0003 & 0.035 \hspace*{\fill} *& 1.307 \\ %
        \rowcolor{lightgray!30} 
        guilt/shame 
        & \textbf{0.0097} & 0.0 \hspace*{\fill} ***& $-$0.139 %
        & 0.0005 & 0.006 \hspace*{\fill} **& 0.640 \\ %
        joy 
        & 0.0065 & 0.0 \hspace*{\fill} ***& 0.173 %
        & 0.0001 & 0.149 & 1.397 \\ %
        pride 
        & 0.0003 & 0.091 & 0.037 %
        & 0.0003 & 0.042 \hspace*{\fill} * & 1.365 \\ %
        relief 
        & 0.0008 & 0.023 \hspace*{\fill} *& 0.063 %
        & 0.0005 & 0.007\hspace*{\fill} * & 1.749 \\ %
        sadness 
        & 0.0007 & 0.031 \hspace*{\fill} *& 0.044 %
        & 0.0000 & 0.470 & 1.138 \\ %
        trust 
        & 0.0067 & 0.0 \hspace*{\fill}***& 0.140 %
        & 0.0000 & 0.654 & 0.886 \\ %
        \hline
        \rowcolor{lightgray!30} 
        \# hedges 
        & 0.0027 & 0.0\hspace*{\fill}*** & $-$0.011 %
        & \textbf{0.0106} & 0.0 \hspace*{\fill}*** & 1.030 \\
        \bottomrule
    \end{tabular}
    \caption{Individual regression results including the explained variance (adjusted $r^2$), respectively, pseudo-$r^2$ for logistic regression, the $p$-value and significance of the effect (***: $p<0.001$, **: $p<0.01$, *: $p<0.05$) and the coefficient, respectively, the logistic odds.}
    \label{tab:single_regression}
\end{table*}
\paragraph{Training Setup} In line with the setup for the \textit{storytelling} feature, we employ an ensemble consisting of RoBERTa classifiers \citep{RoBERTa} fine-tuned on a 10-fold data split and aggregate the predictions into a majority vote. The dataset is originally single-label, with 550 event descriptions generated separately for one emotion. For our target data, we cannot assume a single-label distribution. Thus, we trained a separate classifier for each emotion and downsampled 1650 instances from all other emotion instances for a balanced training set with diverse negative instances.\footnote{The full dataset would result in 8\% positive instances.} Similar to the \textit{storytelling} annotation, we compare two strategies for cross-domain robustness: the event descriptions are available in their original form as well as with salient emotion terms masked. We trained models on both versions to compare the impact of harnessing lexical surface features (\textit{original}) with that of learning more implicit emotion representations (\textit{masked}) and thus gaining more robust performance. %
As the arguments in \cmv are longer than both the texts in the training data and the model's cutoff token length, we additionally split these instances in half and then aggregate the annotations for both halves. 
\paragraph{Results} As the test performance from the training process shows, using \textit{masked} training data improves classification performance significantly (avg. \F increase: 0.074) and exceeds the benchmark performance reported by \citet{troiano2023-emodimensions}. 
The resulting label distribution of the best ensemble is reported in Tab.~\ref{tab:anno-dist}. Apart from \textit{anger} and \textit{disgust}, which occur in almost half of all instances, the data -- especially \ibmargq -- emotions are very sparse, with a ratio of positive instances below 10\% for all other emotions and \textit{surprise} missing entirely from \ibmargq. Thus, we can observe a higher use of emotions in the more subjective \cmv data, together with a general skew towards `indignation-adjacent' emotions like \textit{anger} and \textit{disgust}. While argument-specific emotion use is further analyzed later on (see Sec.~\ref{sec:discussion}), at this point, we observe that very low performance might be related to disuse in argumentation: arguments might intuitively stem from anger or appeal to pride, though arguing from a point of boredom or surprise (our two worst results) might be unusual.

Thus, we continue with annotations from the \textit{masked} and \textit{masked-aggregated} classifiers for our analysis, discarding \textit{surprise} due to its absence in \ibmargq and replacing the binary annotation by averaged classification probabilities in further experiments. We thereby combat data sparseness and leverage prediction confidence (to have indications of `weaker' or `stronger' signs of emotion), making sure that the statistical model can account for robustness.
\subsection{Hedging}\label{sec:data-hedging}
As a surface-level feature, hedges can be extracted through a simple lexicon matching approach. We adapt and combine multiple lexicons from approaches outside the argument domain \citep{islam-etal-hedging-lexicon,sanchez-vogel-2015-hedging,ulinski-hirschberg-2019-crowdsourced}, which are targeted toward similar semi-formal domains (i.e., internet forums) and thus include domain-specific abbreviations and colloquialisms like \textit{AFAIK} (\textit{As far as I know}). Our pipeline first tokenizes and parses the arguments, then matches tokens to a hedging lexicon and further disambiguates terms with simple syntax rules, an example of which can be found in table \ref{tab:hedge-rules}. We were thus able to obtain the number of hedges per argument and create different feature variants, i.e., the overall number of hedges in the first and last sentence, versus in the whole argument instance, as well as the hedge-token ratio for each absolute variant. By including multiple, relative variants of the feature, we are able to abstract from the difference in instance length between the two corpora and accurately portray differences in the usage of hedges. 
Overall, our automated annotation approach proves successful, with increased robustness stemming from our generalization strategies: we find that mixed domain training, masking superficial lexical cues, and employing a deep ensemble is helpful. Although the performance on the argument data is expectably lower than in the training domain, it is nonetheless sufficient for our subsequent analysis and must be seen in relation to the very sparse label distribution in the argument domain.
\begin{table*}
    \centering\small
    \setlength\tabcolsep{3pt}
    \begin{tabularx}{\linewidth}{|X|}
    \hline
There is a difference between a fear of being killed by a terrorist (very small likelihood) and the fear of being *terrorized*. %
I was in Boston when the marathon bombings happened. Terrorism affected everyone on the streets, even though only 3 people were killed. %
The scope of an act of terrorism is much greater than the strict number of casualties. It has a psychological and traumatizing effect on people even in its periphery. %
That being said I am much more afraid of police than an act of terrorism. This is because after the bombings, when Tsarnaev was hiding in a boat about a quarter mile from my apartment at the time, militarized police with bomb dogs searched my house without announcing themselves, came to my door with assault rifles, and kept me locked in my house for a whole day while bomb vans and squad cars raced up and down my street. It was one of the most terrifying days of my life. \textbf{I felt} more electric fear answering the door to what looked like a 9-man SWAT team in full tactical gear and AK-47s than I did in the several previous days of news coverage following the bombing.\\
I don't \textbf{necessarily} agree that having other things to be afraid of, like the abuse of power by the police, makes being afraid of things like acts of terrorism (which are designed to frighten) unreasonable. Fear is real and you don't always have a choice in the matter when it comes to whether or not it will infiltrate your life. \hspace*{\fill}($\Delta0$, \textit{fear, storytelling}, $\varnothing$\textit{hedges}=0.007)\\\hline
Don't mean to be harsh, but that thinking is very dumb. There's a fine line between eating other animals, and cannibalism. Cannibalism is morally wrong because you are practical eating yourself.\hspace*{\fill}($\Delta0$, \textit{guilt/shame, disgust}, $\varnothing$\textit{hedges}=0.0)\\\hline
\textit{Social media brings more good than harm.} Social media helps reconnect with past friends. I was able to reconnect with a childhood best friend not seen in years shortly before he died. For that I am grateful. \hspace*{\fill}(\textit{score}=0.6, \textit{joy, sadness, storytelling}, $\varnothing$\textit{hedge}=0.0)\\\hline
\textit{Social media brings more harm than good.} facts are not checked on social media platforms, allowing public shaming of different figures, hurting them and their career immensely even without them doing anything wrong \hspace*{\fill}(\textit{score}=0.47, \textit{disgust, anger}, $\varnothing$\textit{hedge}=0.0)\\\hline
\textit{Gambling should be banned.} Gambling \textbf{can} be addictive and those who become addicted face severe financial and personal consequences such as bankruptcy, jail (from financial crimes as stealing or embezzlement to support the addiction),
divorce and suicide. \hspace*{\fill}(\textit{score}=1.0, \textit{fear, sadness}, $\varnothing$\textit{hedge}=0.11)\\\hline
\textit{Flu vaccination should not be mandatory.} While \textbf{I believe} that flu vaccines are beneficial to people, I do not \textbf{believe} they \textbf{should} be mandatory because I \textbf{should} have a right to decide if I want to take a risk with my health.\hspace*{\fill}(\textit{score}=0.8, $\varnothing$\textit{hedge}=0.12)
\\\hline
    \end{tabularx}
    \caption{Fully annotated examples from \cmv and \ibmargq, with all positive labels listed below the post text and hedge terms rendered bold.}
    \label{tab:feature_ex}
\end{table*}

\section{Regression Analysis}\label{sec:analysis}

Following the successful automated annotation procedure, we implement a regression analysis to investigate the impact of all 16 features (1 \textit{storytelling}, 9 emotions excluding \textit{boredom, surprise}, 6 \textit{hedging}) as independent variables on the dependent variables of \textit{quality} score in \ibmargq and \textit{persuasion} label in \cmv. We use the Python \textit{statsmodels} library \citep{seabold2010statsmodels} to implement OLS linear regression with $t$-testing for significance on the \textit{quality} score of \ibmargq and logistic regression with $z$-testing for significance on the binary \textit{persuasiveness} label of \cmv. 
To measure how much variance can be explained by individual features and how much additional variance can be explained by combining features, we compare regression models that employ a single feature as IV to richer models with multiple IVs and two-way interactions.

\begin{figure}[t]
\centering
    \includegraphics[width=\linewidth]{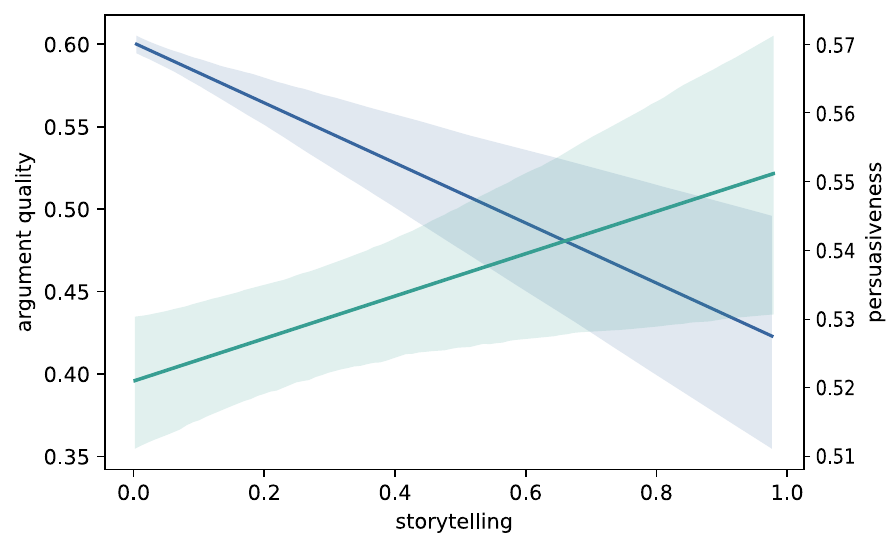}
    \caption{\textit{Storytelling} effect on \ibmargq argument quality (teal, left $y$-axis) and on \cmv persuasion (blue, right $y$-axis), with confidence intervals.}
    \label{fig:story_regression}
\end{figure}

In comparing individual regression results of all features (see Tab.~\ref{tab:single_regression}), two major divergences between the two corpora emerge. Firstly, both \textit{storytelling} ($\beta=-.182$) and the absolute \textit{hedging} count ($\beta=-.011$, for hedging in all variants, see Appendix Tab.~\ref{tab:hedge-lr-ibm}) are highly significant negative predictors of argument quality in \ibmargq, but significantly improve persuasiveness in \cmv ($\beta_{story}=0.138$, $\beta_{hedge})=0.030$, cf. Fig.~\ref{fig:story_regression}), with \textit{hedging} constituting the most informative feature for this dataset. Secondly, an overall trend of greater and more frequent significant effects can be observed for \ibmargq argument quality than for \cmv persuasion. This trend comes along with a greater predictive power of the \ibmargq models,\footnote{While the adjusted $r^2$ of the \ibmargq models can be interpreted as the percentage of explained variance, this cannot be compared directly to the pseudo-$r^2$ of the logistic \cmv models. The general difference in magnitude nonetheless holds.} and is continued in the best multiple regression model, which includes more IVs for \ibmargq than for \cmv.

In contrast to these domain differences, the impact of emotions on argument strength is largely domain-independent, with direction and magnitude of effects comparable between \ibmargq and \cmv for all emotions but \textit{trust}. As such, the emotions with the highest impact on argument strength are \textit{guilt/shame} and \textit{disgust}, which both significantly decrease argument strength. For these emotions, as for most others, emotion polarity matches effect direction, including the significant emotions of \textit{relief} (both corpora), \textit{pride} (\cmv), and \textit{joy} (\ibmargq). Two emotions contradict this trend: opposite to their polarity, \textit{fear} ($***$ \ibmargq; $*$ \cmv) and \textit{sadness} ($*$ \ibmargq) improve argument strength in both corpora.

To further investigate the interplay between different argument features, we implemented two multiple regression analyses with and without interaction. We used stepwise multiple regression, where individual IVs or two-way feature interactions are added incrementally according to their AIC value (predictive improvement relative to model size), while ensuring the significance of added IVs compared to the smaller model through ANOVA (\ibmargq) and F-test (\cmv).

\begin{figure}[t]
\centering
    \includegraphics[width=\linewidth]{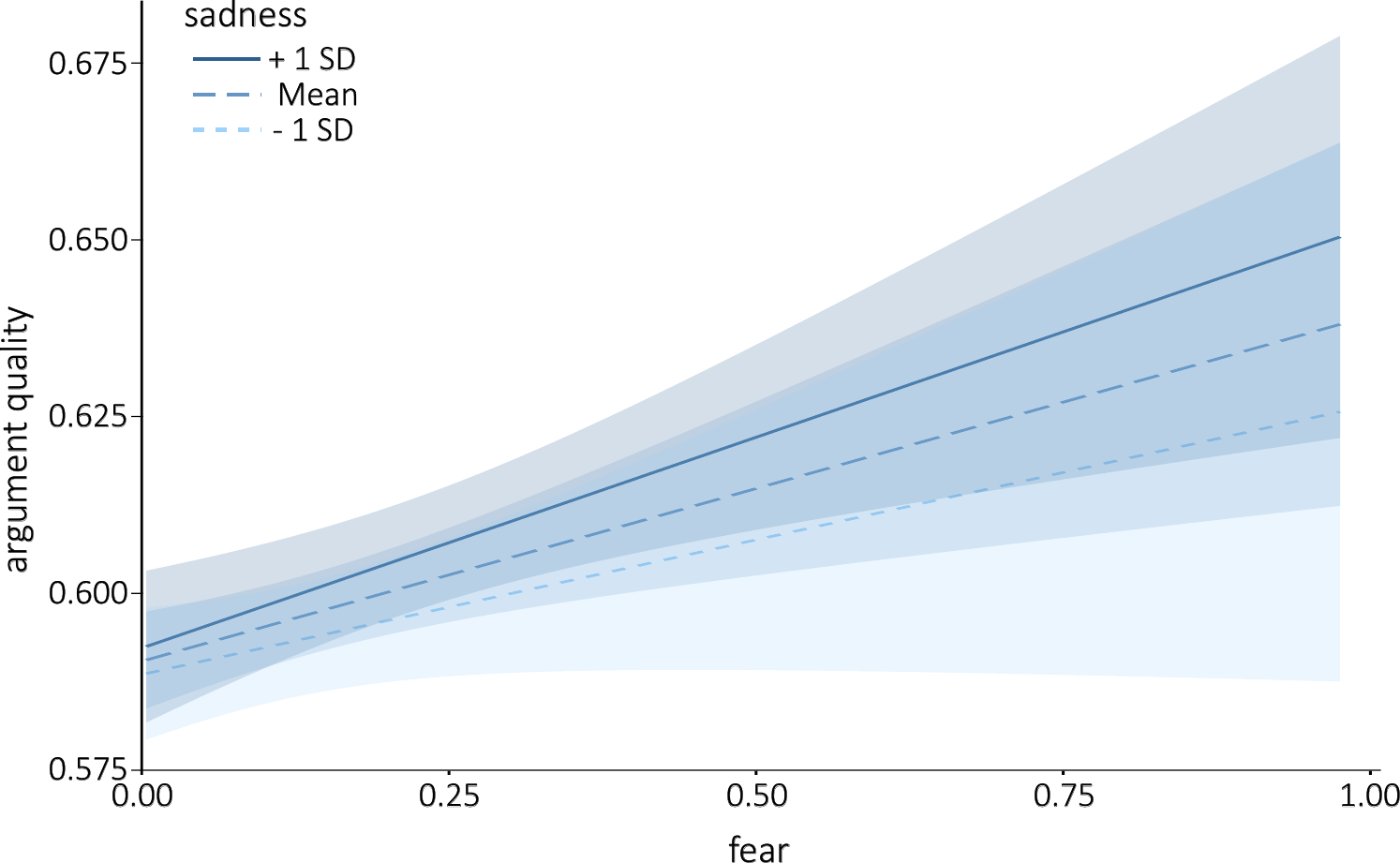}
    \caption{Interaction between \textit{fear} ($x$-axis) and \textit{sadness} (standard deviations shown through hue and dashing) on \ibmargq with confidence intervals.}
    \label{fig:fear_sadness}
\end{figure}

The full models reveal the consistency of most effects on argument strength, as the most informative features of \textit{guilt/shame} retain their salience, and notable observations like the diverging effect of \textit{storytelling} on persuasion vs. quality are present in the full model as well. Interactions show a general trend of same-directed features combining to an effect of greater magnitude, as seen with the individually positive features of \textit{fear} and \textit{sadness} interacting on \ibmargq argument quality to form a highly positive combined effect while their individual effects are neutralized (Fig.~\ref{fig:fear_sadness}). The full models with interaction further show the persistant importance of \textit{storytelling}, which (in contrast to the individual \ibmargq regression) has a positive effect in both datasets. The final explained variance is 3.96\% adjusted $r^2$ for \ibmargq and 1.36\%  pseudo-$r^2$ for \cmv. Although generally low, these values are reasonable and expected for a regression on the complex notion of argument strength, considering the exclusion of contextual information (e.g., topic, demographics of the annotators/OPs) and overall low values (and thus error margins) for both independent and target variables (between 0 and 1). 

\paragraph{Discussion}\label{sec:discussion}
The diverging effects of \textit{hedging} and \textit{storytelling} show the importance of domain-aware rhetoric: harnessing such subjective features significantly improves the odds of subjective persuasion, but in the objective domain of \ibmargq, they hinder argumentative success (cf. Fig. \ref{fig:story_regression}). As all subjective features are infrequent in \ibmargq, where arguments were mandated as short and impersonal, their successful use in \cmv seems intuitive, indicating their importance for non-experts.

When viewing the results of our two steps side by side, it is apparent that emotions are utilized differently in argumentation than in their original context. While \textit{disgust} and \textit{anger} are overrepresented compared to all other features, a qualitative analysis (see Tab.~\ref{tab:feature_ex}) shows their idiosyncratic appearance in arguments. Both emotions seem closer to indignation or `righteous' anger, and occur, with the similarly impactful \textit{guilt/shame}, almost always explicitly targeted towards either another participant (`\textit{that thinking is very dumb}') or the topic under discussion (`\textit{allowing public shaming}'). The very beneficial emotions of \textit{fear} and \textit{sadness}, on the other hand, seem reframed as an appeal to universal concerns instead of individual experiences, even when combined with personal experiences: `\textit{personal consequences}', `\textit{whether or not it will infiltrate your life}'. Therefore, we hypothesize that discrete emotions are utilized in two diverging strategies of \textit{emotional attacks} and \textit{emotional appeals}. While the latter are highly efficient in persuasion, the former hinder argument strength but are much more frequent in the data. 

\section{Conclusion}
This paper has investigated the impact of a number of subjective features on two diverging facets of argument strength. To that end, we first determined the feasibility of large-scale automated annotation of our subjective features, to then systematically reveal correlations through a regression analysis. 
We could reveal a significant effect of almost all observed features on argument strength, thus affirming \textbf{RQ1}. We moreover demonstrated the importance of argument context for subjective features, as personal anecdotes and uncertainty indicate a lack of rhetoric proficiency in objective settings, but strengthen arguments in the subjective domain, thereby affirming \textbf{RQ2}. Further qualitative assessment shows frequent \textit{emotional attacks} with righteous indignation impeding argument strength, while less frequent \textit{emotional appeals} to empathy and universal fears seem to strengthen arguments. This finding reveals an avenue for continuing argument-specific emotion research, a research gap that is further emphasized by the results of our automated modeling. We could successfully model \textit{storytelling} and most emotions automatically due to our robustness strategies of employing a deep ensemble based on training data from mixed-domains and with masked surface lexical cues. Thus, in situations where large-scale gold data is neither available nor easily attainable, these strategies constitute an acceptable alternative. However, the unbalanced and idiosyncratic distribution of emotions also reveals the limits of cross-domain approaches, as some emotions are used extremely seldomly, or appear changed from their original definition. We thus highlight once more the need for emotion data and definitions directed at argumentation, a research gap that has recently been addressed for German text by \citet{greschner2024fearfulfalconsangryllamas} and should receive further attention on a larger scale.

\section*{Acknowledgments}
 We would like to thank the anonymous reviewers whose feedback helped us improve this paper. This research has been partially funded by the Bundesministerium für Bildung und Forschung (BMBF) through the project E-DELIB (Powering up e-deliberation: towards AI-supported moderation)

\section*{Limitations}

Apart from the obvious constraint of English-only modeling, automatically annotating the independent variables bears the risk of modeling the influence of features that differ from the named features. For the features of storytelling and hedging, our success in recreating results from existing works leads us to believe that the annotations are acceptable even on unseen data. For our emotion features, we rely on our strategies of masking salient surface features and aggregating predictions for long instances to lead to an acceptable performance based on the good results on the heldout training data.
Thus, we believe our regression to realistically model the influence of the remaining investigated features. 
This influence is very small, as denoted by the low $r^2$ and pseudo-$r^2$ scores of the regression models. However, while this shows that the features investigated here cannot fully explain argument strength, the high significance of most features nonetheless shows their importance for argument strength. As previous research shows, argument strength is a complex and subjective feature. We thus expect that a model regressing argument strength to a higher degree must include context, such as prior beliefs and demographic features of the annotators/OP and the author, topic information, or discussion history. The significance of our results constitutes one step in a growing field of research aiming to explore argument strength as a multi-faceted complex feature.

\section*{Ethical Considerations}
As always in the analysis of argument strength, our results may potentially be exploited in the persuasion strategies of bad actors. However, we observed significant but very small effects that may be less impactful than demographic and contextual features, which we omitted. Further, features like emotions or uncertainty are likely used intuitively and, as shown elsewhere \citep[cf. e.g.,][]{vasilieva2004gender}, used differently depending on demographic factors. While reporting negative influence might discredit argument strategies used by already disadvantaged groups, we believe that our features bear no inherent demographic inclination and understanding such effects is the first step to encourage thoughtful argumentation.

\bibliography{custom}

\appendix

\section{Supplementary Material}\label{sec:appendix}
\subsection{Data and Annotation}
Table \ref{tab:argument-example} shows exemplary instances of the base corpora we used, while tables \ref{tab:test-eval} and \ref{tab:hedge-rules} explicate the further annotation process.
\subsection{Regression Results}
Reported below are the regression results for all hedging variants (Tab.~\ref{tab:hedge-lr-ibm}), the results of full stepwise regression model with interaction (Tab.~\ref{tab:full_regression}), and two exemplary regression plots (Fig.~\ref{fig:story_regression}, \ref{fig:fear_sadness}).%

\begin{table}[!h]
    \centering\small
    \setlength\tabcolsep{3pt}
    \begin{tabular}{lccc}
\toprule
& \multicolumn{2}{c}{\textbf{Training variant}} & \textbf{Benchmark}\\
\textbf{Feature}
& \textit{masked}
& \textit{orig}\\
\hline
\textit{anger} %
& 0.76($\pm$0.03)
 & 0.69($\pm$0.04) 
 & 0.53 \\
\textit{boredom} %
& 0.88($\pm$0.03) 
 & 0.84($\pm$0.02)  
 & 0.84 \\
\textit{disgust} %
& 0.82($\pm$0.03) 
 & 0.75($\pm$0.04) 
 & 0.66 \\
\textit{fear} %
& 0.81($\pm$0.03) 
 & 0.72($\pm$0.03) 
 & 0.65 \\
\textit{guilt/shame} %
& 0.85($\pm$0.03) 
 & 0.80($\pm$0.02) 
 & 0.48/0.51 \\
\textit{joy} %
& 0.77($\pm$0.03) 
& 0.71($\pm$0.02) 
 & 0.45 \\
\textit{pride} %
& 0.83($\pm$0.03) 
 & 0.75($\pm$0.03) 
 & 0.54 \\
\textit{relief} %
& 0.82($\pm$0.03) 
 & 0.70($\pm$0.25) 
 & 0.63 \\
\textit{sadness} %
& 0.81($\pm$0.04)
 & 0.73($\pm$0.03) 
 & 0.59 \\
\textit{surprise} %
& 0.78($\pm$0.02)
 & 0.67($\pm$0.04) 
 & 0.53 \\
\textit{trust} %
& 0.85($\pm$0.02) 
& 0.80($\pm$0.02) 
 & 0.74 \\ 
\midrule
& \textit{mixed}
& \textit{one} & \\
 \hline
\textit{storytelling} %
& 0.82($\pm$0.03) 
& 0.78($\pm$0.05) 
 & 0.76-0.94\\ 
\bottomrule
\end{tabular}
    \caption{\F performance of the ensemble classifiers on the heldout test set of their respective training data with standard deviance reported in brackets. The last column lists the originally reported benchmark: \citeauthor{troiano2023-emodimensions}'s (\citeyear{troiano2023-emodimensions} text-based classifier (multilabel versus our single label classifiers) and the best overall approach by \citet[performance is reported separately for three subsets, thus ranging between values]{falk-lapesa-storytelling}.}
    \label{tab:test-eval}
\end{table}

\begin{table}[!h]
    \centering{\small
    \setlength\tabcolsep{3pt}
    \begin{tabularx}{\linewidth}{p{.15\linewidth}X}
        \toprule
        \textbf{Term} & \textbf{Rule}\\\hline
        \textit{about}, \textit{around} & If the token is an adjective, it is a non-hedge.\newline
        \textbf{Hedge:} There are \textit{around} 10 million packages in transit right now.\newline
        \textbf{Non-hedge:} We need to talk \textit{about} Mark.\\\hline
        \textit{pretty} & If the token is used as adverbially, it is a hedge.\newline
        \textbf{Hedge:} I am \textit{pretty} certain about this statistic.\newline
        \textbf{Non-hedge:} She has a really \textit{pretty} cat.\\\hline
        \textit{impression} & If the token has a 1. person possessive pronoun as dependent or its head has a 1. person nominal subject as a second dependent, it is a hedge.\newline
        \textbf{Hedge:} I get the \textit{impression} that we have to wait longer for official information.\newline
        \textbf{Non-hedge:} The protagonist's performance left a lasting \textit{impression} on everyone.\\\bottomrule
    \end{tabularx}}
    \caption{Examplary hedge disambiguation rules, the first of which is lifted from \citet{islam-etal-hedging-lexicon}.}\label{tab:hedge-rules}
\end{table}

\begin{table*}[!h]
   \centering\small
   \setlength\tabcolsep{3pt}
   \begin{tabular}{p{0.53\textwidth}p{0.43\textwidth}}
    \toprule
    \textbf{\ibmargq}\\
    \textit{We should ban fossil fuels.}
    fossil fuels are bad for the country because of your country dont have them they have to be in an inferior position to ather countrys. \hfill(\texttt{score=0.18})
    & \textit{We should ban fossil fuels.}
    Fossil fuels destabilize the ecosystem which will harm future generations. \hspace*{\fill}(\texttt{score=1.0})\\
    \hline    
        \textbf{\cmv}\\
        \multicolumn{2}{p{0.98\textwidth}}{\textit{CMV: Driving a car is insanely risky and probably the most dangerous thing you do in your everyday life.}
        I find it difficult to understand how so many people enjoy driving a car or can even relax while doing it. I am almost continually tense while on the road thinking about what's at stake (and I've been driving for almost 20 years). {[}\dots{]} 
        }\\\hline%
        By the death rate, eating unhealthy is the most dangerous thing that you can do. Cellular reproduction is up there are well. Then there's realizing your worthless and life is futile, then taking your own life. \newline
        Looking at the CDC, suicide isn't on there. But breathing shit other than oxygen and nitrogen is up there. So is, the fatty food thing again. \hspace*{\fill}($\Delta0$)
        & 
        Mortality for drivers in the US is roughly 50 per millions. Death while working in construction in 2006 was 108 per millions. Driving is not the most dangerous thing these workers do in their everyday life. (edit. The more i'm looking into it the more I find that stats regarding this subject varies a lot.) \hspace*{\fill}($\Delta1$)\\
        \bottomrule
    \end{tabular}
    \caption{Examples from \ibmargq and \cmv of a bad (left) and good (right) argument about the same topic, with the shortened original post from \cmv given above the two answering arguments.}\label{tab:argument-example}
\end{table*}

\begin{table*}[ht]
    \centering\small
    \begin{subtable}{.48\textwidth}
    \begin{tabular}{llrrll}
        \textbf{score} & \textbf{sent} & $\mathbf{r^2}$ & \textbf{Coef} & \textbf{\textit{p}}\\\hline
        \textit{absolute} 
        & \textit{first} & \textbf{0.0044} & $-$0.029 & \textbf{0.0} & *** \\
        & \textit{final} & $-$0.0002 & 0.001 & 0.894 \\
        & \textit{all} & 0.0027 & $-$0.011 & 0.0 & *** \\
        \textit{ratio} 
        & \textit{first} & 0.0036 & $-$0.160 & 0.0 & *** \\
        & \textit{final} & 0.0007 & $-$0.159 & 0.026 & *\\
        & \textit{all} & 0.0036 & $-$0.296 & 0.0 & *** \\
        \bottomrule
    \end{tabular}
    \caption{\ibmargq}
    \end{subtable}
    \begin{subtable}{.48\textwidth}
        \begin{tabular}{lllrll}
        \textbf{score} & \textbf{sent} & \textbf{pseudo-}$\mathbf{r^2}$ & \textbf{Odds} & \textbf{\textit{p}}\\\hline
        \textit{absolute} & \textit{first} & 0.00005 & 1.018 & 0.358 \\
        & \textit{final} & 0.0 & 0.999 & 0.947 \\
        & \textit{all} & \textbf{0.01056} & 1.030 & \textbf{0.0} & *** \\
        \textit{ratio} & \textit{first} & 0.00002 & 1.235 & 0.565 \\
        & \textit{final} & 0.00012 & 0.579 & 0.174 \\
        & \textit{all} & 0.00035 & 0.124 & 0.018 & * \\
        \bottomrule
        \end{tabular}
        \caption{\cmv}
    \end{subtable}
    \caption{Individual regression results of each hedging variant as IV on \ibmargq argument quality and \cmv persuasiveness. The variants are listed by \textbf{score} (absolute or ratio values) and the \textbf{sent}ence for which the score is calculated. Reported are the adjusted \textit{r}\textsuperscript{2} percentage, respectively, pseudo-$r^2$ for logistic regression, the coefficient/odds of the feature variant and the effect's $p$-value/significance.}\label{tab:hedge-lr-ibm}
\end{table*}

\begin{table*}
    \centering\small
    \begin{subtable}[b]{.48\textwidth}
    \centering
\begin{tabular}{lrl}
\toprule
\textbf{IVs} & \textbf{adjusted \textit{r}\textsuperscript{2}} & \textbf{sign.} \\
\midrule
guilt/shame & 0.971 & x \\
$+$ \textit{all hedge}$\times$\textit{storytelling} & 1.723 & *** \\
$+$ \textit{fear}$\times$\textit{guilt/shame} & 2.273 & *** \\
$+$ \textit{joy} & 2.602 & *** \\
$+$ \textit{disgust}$\times$\textit{sadness} & 3.082 & *** \\
$+$ \textit{boredom}$\times$\textit{pride} & 3.484 & *** \\
$+$ \textit{pride}$\times$\textit{relief} & 3.579 & * \\
$+$ \textit{pride}$\times$\textit{sadness} & 3.715 & ** \\
$+$ \textit{disgust}$\times$\textit{fear} & 3.774 & * \\
$+$ \textit{sadness} & 3.845 & * \\
$+$ \textit{storytelling} & 3.904 & * \\
$+$ \textit{fear}$\times$\textit{relief} & 3.962 & * \\
\bottomrule
\end{tabular}
\caption{\ibmargq}
    \end{subtable}
    \begin{subtable}[b]{.48\textwidth}
    \centering
    \begin{tabular}{lrl}
\toprule
\textbf{IVs} & \textbf{pseudo-\textit{r}\textsuperscript{2}} & \textbf{sign.} \\
\midrule
\textit{\# hedge} & 0.0106 & x \\
$+$ \textit{disgust$\times$guilt/shame} & 0.0113 & *** \\
$+$ \textit{fear}$\times$\textit{pride} & 0.0119 & ** \\
$+$ \textit{anger}$\times$\textit{relief} & 0.0123 & ** \\
$+$ \# \textit{hedge}$\times$\textit{anger} & 0.0128 & ** \\
$+$ \textit{disgust}$\times$\textit{pride} & 0.0132 & ** \\
$+$ \# \textit{hedge}$\times$\textit{guilt/shame} & 0.0136 & * \\
\bottomrule
\end{tabular}
    \caption{\cmv}
    \end{subtable}
    \caption{Features and explained variance of the interactive multiple regression on \ibmargq and \cmv. The model is built stepwise by adding features/interactions with the highest AIC (Akaike Information Criterion relating predictive power to model size) and stops if no improvement is observed. The significance (***: $p<0.001$, **: $p<0.01$, *: $p<0.05$) of adding each new feature is tested via ANOVA for \ibmargq and via F-test for \cmv. %
    }\label{tab:full_regression}
\end{table*}
\end{document}